\documentclass{article}
\usepackage{ijcai23}

\usepackage{times}
\usepackage{soul}
\usepackage{url}
\usepackage[hidelinks]{hyperref}
\usepackage[utf8]{inputenc}
\usepackage[small]{caption}
\usepackage{graphicx}
\usepackage{amsmath}
\usepackage{amsthm}
\usepackage{booktabs}
\usepackage{algorithm}
\usepackage{algorithmic}
\usepackage[switch]{lineno}
\usepackage{amsmath}
\usepackage{amsthm}
\usepackage{amssymb}
\usepackage{booktabs}
\usepackage{algorithm}
\usepackage{algorithmic}

\newtheorem{assumption}{Assumption}

\title{Rethinking Precision of Pseudo Label:\\
Test-Time Adaptation via Complementary Learning}

\author{
Jiayi Han$^{1\dag}$
\and
Longbin Zeng$^{1\dag}$\and
Liang Du$^{2}$\and
Weiyang Ding$^1$\And
Jianfeng Feng$^1$
\thanks{$^{\dag}$These authors contribute equally.}
\affiliations
$^1$Fudan University\\
$^2$Tentcent Inc.\\
\emails
\{dingwy,jffeng\}@fudan.edu.cn,
}

\begin{document}

\maketitle

\begin{abstract}
In this work, we propose a novel complementary learning approach to enhance test-time adaptation (TTA), which has been proven to exhibit good performance on testing data with a different distribution. 
In test-time adaptation tasks, information from the source domain is typically unavailable and the model has to be optimized without supervision for test-time samples.
Hence, usual methods assign labels for unannotated data with the prediction by a well-trained source model in an unsupervised learning framework. 
Previous studies have employed unsupervised objectives, such as the entropy of model predictions, as optimization targets to effectively learn features for test-time samples.
However, the performance of the model is easily compromised by the quality of pseudo-labels, since inaccuracies in pseudo-labels introduce noise to the model.
Therefore, we propose to leverage the "less probable categories" to decrease the risk of incorrect pseudo-labeling.
The complementary label is introduced to designate these categories. We highlight that the risk function of complementary labels agrees with their Vanilla loss formula under the conventional true label distribution.
Experiments show that the proposed learning algorithm achieves state-of-the-art performance on different datasets and experiment settings.
\end{abstract}

\section{Introduction}
\label{sec:intro}

Deep-learning techniques have demonstrated exceptional performance when trained and evaluated on data from the same distribution. Nevertheless, this performance may not generalize well to unseen data with distribution shifts, for instance, image corruption. 
However, generalization to diverse data shifts is restricted due to the infeasibility of incorporating a sufficient number of augmentations during training to account for the wide range of possible data shifts \cite{mintun2021interaction}.
An effective technique to transfer the model to a new related data domain is required, known as domain adaptation. In this work, we focus on the problem of fully test-time adaptation (TTA), where the source data is not available during adapting to unlabeled test data. We only access the trained model in the source domain and update the parameters via a few optimization steps on an unsupervised objective involving the test samples from the target distribution. 
\par
Different works exist to improve the model's performance during the testing procedure. \cite{liang2020we} learns the target-specific feature extraction by exploiting both information maximization and self-supervised pseudo-labeling to implicitly align representations from the target domains to the source. \cite{wang2020tent} reuse the trained model to represent the label distribution of testing data and propose to minimize the entropy loss to maximize the prediction confidence during adaptation. \cite{mummadi2021test} extend the work of Wang~\cite{wang2020tent} by introducing a novel loss function and prepending a parameterized input transformation module, effectively improving the robustness. \cite{zhang2021understanding} demonstrates that deep models are capable of converging to incorrect labels. Therefore, the approaches mentioned above have a common drawback of requiring a sufficient reliable pseudo label of unseen target testing data. 

\begin{figure}[tbp]
    \centering
    \includegraphics[width=.9\linewidth]{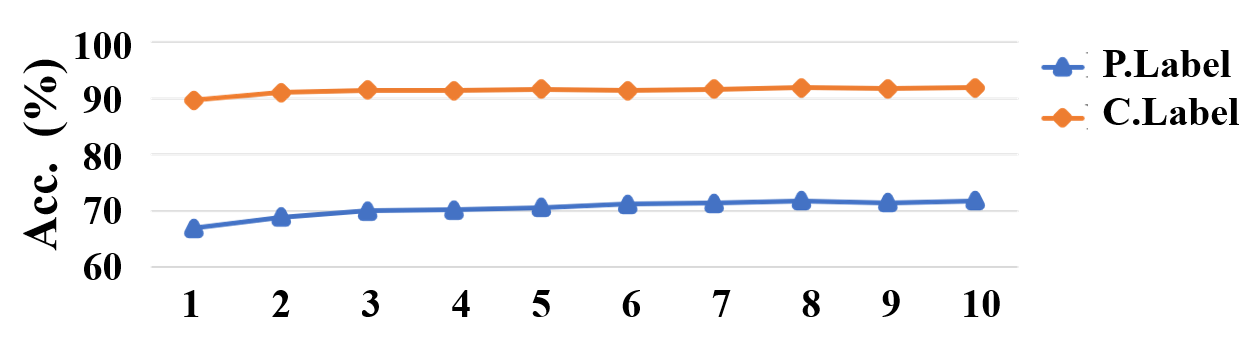}
    \caption{The accuracy of the positive pseudo labels and complementary ones in testing-time adaptation. The negative ones make fewer mistakes in the prediction of correct labels. P.Label and C.Label represent the pseudo label and complementary label, respectively.}
    \label{fig: Motivition}
\end{figure}

Given a decision function in multi-class classification, identifying a class label that is incorrect for the new coming instance is often much convinced than identifying the actual label. As shown in Fig.~\ref{fig: Motivition}, the predictions of complementary labels are much more accurate than the naive (positive) pseudo labels. 

\begin{figure}[htbp]
    \centering
    \includegraphics[width=1.\linewidth]{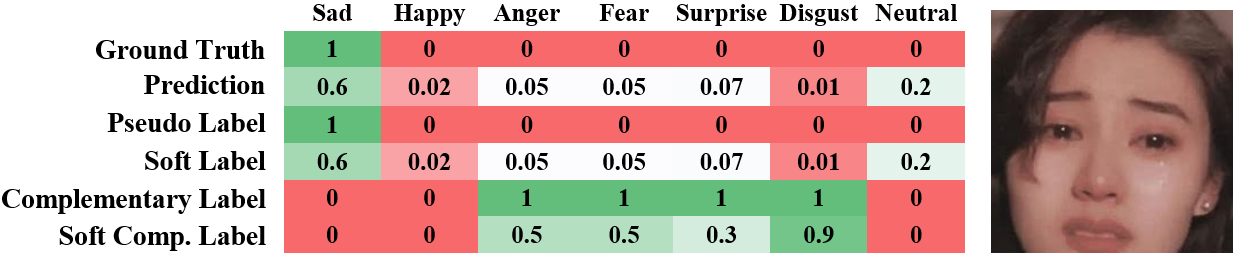}
    \caption{An illustration of different types of labels. The right picture is the input sample. ``Prediction'' represents the predicted probability distribution. Pseudo label, soft label, complementary label, and soft complementary label are generated accordingly.
    }
    \label{fig:cl}
\end{figure}

The label that figures out the categories that the sample does not belong to is called a complementary label. In Fig.~\ref{fig:cl}, we give an example of a Complementary label.

Complementary labels carry useful information and are validated in comprehensive experiments on several benchmark datasets \cite{yu2018learning}. We could mitigate the above-mentioned error caused by incorrect annotating by introducing a novel learning strategy with complementary labels instead of directly assigning a pseudo label to take the place of the inaccessible label. Complementary labels are easily obtainable, especially in the TTA we can reuse the source model to specify the least probable categories. Optimizing the model by tuning the decision function to suppress the less possible complementary labels makes sense. In the context of the forward decision problem, complementary labels provide limited information, however, they significantly mitigate the negative effects of high-confidence false tags on the model.
\par 
Motivated by the pseudo-labeling method in Tent \cite{wang2020tent}, we here model the learning procedure of annotating complementary labels via probabilities. In this work, we propose a complementary learning (CL) framework in TTA without any knowledge of source data. Specifically, we normalize the predicted distribution of the trained source model and then assign the complementary distribution. Then, we modify standard loss functions proposed for learning with actual labels so that the modifications can be employed to learn with complementary labels efficiently. We highlight that this proposed risk function agrees with vanilla learning with ordinary labels and rapidly converges to an optimal one. Moreover, we also empirically demonstrate its effectiveness by exploiting examples in fully test-time adaptation problems. The main contributions of this paper are listed as follows:
\begin{enumerate}
    \item To the best of our knowledge, this is the first work that proposes to utilize the complementary labels on the TTA task.
    \item The proposed risk function theoretically ensures that the classifier learned with complementary labels converges to an optimal hypothesis.
    \item Our proposed algorithm reaches state-of-the-art performance on different datasets and experiment settings. Extensive evaluations demonstrate the effectiveness of the proposed method.
\end{enumerate}
\section{Related Work}

\subsection{Model Training with Noisy Labels}
Training model with noisy labels is an important issue in deep-learning tasks, as incorrect labels are inevitable, and models have the ability to overfit on noisy data \cite{zhang2021understanding}. A basic approach is to purify the labels. \cite{wu2020topological} proposes to filter the clean samples via their topological information in the latent space. Similarly, \cite{kim2021fine} proposes to filter the noisy samples according to the eigenvectors. \cite{xia2021sample} assumes that the volume of loss could be a clue of clean data and utilize interval estimation to find the most probable noisy labels. However, directly eliminating the samples with noisy labels reduces the size of the available data. To explicitly utilize the dataset, some approaches propose refurbishing the noisy labels \cite{zheng2020error,wang2020suppressing,chen2021beyond}. These approaches managed to detect the samples with incorrect labels and replace the annotation with the model's prediction. 

\subsection{Unsupervised Domain Adaption (UDA) and TTA}

UDA is a similar task to TTA. UDA aims to improve a model trained on the source domain and validate on the target domain.
There are two main differences between UDA and TTA: First, UDA is generally an offline training strategy that allows collecting the whole target dataset for offline finetuning. Opposite to UDA, TTA can only see the current and the past mini-batches. Second, knowledge of the source domain is allowed to be accessible to UDA models, which is not supposed to be involved in TTA \cite{wang2020tent}.

For UDA, since the training data in the source domain are supposed to be available, many works optimize the model via two tasks: the main classification task supervised by data from the source domain, and the feature alignment task that minimizes the gap of feature distribution of data from both domains \cite{Long_2013_ICCV,chen2019joint}.
\cite{liu2018detach} disentangles the feature representations from the source and target domains with source-domain supervision.

For TTA, a basic approach is to introduce pseudo-labeling. To further involve the confidence, \cite{wang2020tent} suggests using simple Shannon entropy $H=-\sum_{c}p(y_c)log(p(y_c))$ as the target of optimization other than the standard pseudo label. Meanwhile, they should solely optimize the batch normalization layer to avoid the model collapsing.
\cite{wang2022continual} proposes to utilize test-time augmentation to generate more reliable soft pseudo labels and fit the model to the labels by cross-entropy loss. Different from \cite{wang2020tent}, they propose that the environment is constantly evolving and the model must adapt to these changes over time. In this work, we conduct experiments on both settings to demonstrate the effectiveness of the proposed complementary learning. 


\section{Motivation Analysis}
To illustrate our motivation, we start by calculating the accuracy of the pseudo label and complimentary label. Consider a well-trained classifier $f$. Assume $f(x)$ is an ideal estimation of the probability that $x$ belongs to each category. If $f(x)_c<\theta_c$, we assume that $x$ may not belongs to the $c_{th}$ category, so the $c_{th}$ category could be viewed as a negative category. We call the set of negative categories ``complementary label''. Pseudo-labeling assumes that the most probable category is the true category so that when $f(x)$ is an ideal estimation, the probability of the pseudo label being the ground truth is $P=f(x)_{max}$. 
Then we define the accuracy of complementary labels. Since there could be more than one negative category for each sample, if the ground truth category is not involved in the complementary label of $x$, we name it a ``correct complementary label''. The probability $P_{cl}$ of the complementary label being correct could be calculated as
\begin{equation}
    \begin{aligned}
    P_{cl} & = \prod\limits_{c\in \bar{y}_{x}}(1-f(x)_c) \\
      & \geq \prod\limits_{c\in \bar{y}_{x}}(1-\theta_c) \\
      & \geq (1-\Theta)^{\vert \bar{y}_{x} \vert} \\
      & \geq (1-\Theta)^{C-1}.
    \end{aligned}
    \label{equ:pneg}
\end{equation}
in which $C$ is the number of categories, $\bar{y}_x$ represents the set of the negative categories of sample $x$ filtered by the thresholds, and $\Theta$ represents the largest threshold of all categories. According to Equ.~\ref{equ:pneg}, when $\Theta<1-f_{max}^{\frac{1}{C-1}}$, $P_{cl}>P$. That is to say, to a given sample, if the threshold is small enough, the accuracy of the complementary label is higher than it of the pseudo label. 

\section{Methods}

\subsection{A Basic Approach}
A direct approach to utilizing the complementary label is finding the negative categories and minimizing the predicted probability of these selected negative categories. We only minimize the predicted probabilities on the negative categories and ignore the non-negative ones. Specifically, given model $f$, which is well-trained on the source domain, and a batch of data $D=\{x_i\}_{i=1}^{N}$, we optimize the model according to the following objective function:
\begin{equation}
    \begin{array}{l}
        \mathcal{L} = \frac{-1}{N\times C}\sum\limits_{x\in D}\sum\limits_{c=1}^{C}
        \delta(\theta_c > f(\boldsymbol{x})_c)p\operatorname{log}(p),p=1-f(\boldsymbol{x})_c,
    \end{array}
\end{equation}
in which $C$ is the number of categories and $\theta_{c}$ is a chosen threshold to filter the complementary label $c$. The binarized function ``$\delta(\cdot)$'' returns a value of 1 if the enclosed assertion is true and a value of 0 otherwise. 
In the rest of the paper, we call the basic approach {\bf BCL} (basic complementary learning).

\subsection{Learning with Confidence}
If the probability that a sample belongs to a certain category is below a certain threshold, that category is considered the negative category for that sample. However, if the probability is significantly lower than the threshold, the certain category is a negative category with much confidence, which may not have been adequately modeled by the above-mentioned basic method. To address this issue, we enhance the BCL approach by incorporating confidence. We call the enhanced version {\bf ECL} (enhanced complementary learning).
\par
In the framework of learning input-output relationship, the trained data samples $(x_{i}, y_{i})$ (input-output pairs) are drawn from an underlying distribution $p^{*}(x, y)$. A well-trained model could specify a conditional distribution $p_{\theta}(y|x)$ for a given input $x$ with a set of possible parameters $\theta$. The joint distribution of input-output pairs is achieved, i.e, $p_{\theta}(x, y) = p_{\theta}(y|x) p^{*}(x)$, where $p^{*}(x)=\int p^{*}(x, y) dy$ is the marginal distribution of the input. The goal is to minimize the KL divergence between the true distribution $p^{*}(x, y)$ and the model prediction $p_{\theta}(x, y)$, which is equivalent to maximum likelihood, i.e, by minimizing the following the risk function:
\begin{equation}
    \mathcal{R}(\theta) =\langle -log\, p_{\theta}(x, y)\rangle_{p^{*}} =\langle-log\, p_{\theta}(y|x)\rangle_{p^{*}} + \text{const},
\end{equation}
where $\langle\,\cdot \rangle_{p^{*}}$ denotes the expection over true data distribution. In the usual classification case with ordinarily labeled data at hand, approximating the risk empirically is straightforward,
\begin{equation}\label{emperical_risk}
   \mathcal{R}(\theta) = \frac{1}{n} \sum_{i=1}^n -log\, p_{\theta}(y_{i}|x_{i}).
\end{equation}
Here, $n$ is the number of instances in the available dataset, and $y_{i}$ is the given class
label for the $i$-th instance. The equality holds due to the following two facts:1) the conditional distribution of class label $q^{*}(y|x)$ for every training instance is a normalized delta function; 2) the number of data instances in the real world is generally finite.
\par 
Consider in this special case of closed-set test-time adaptation problem, we have already obtained a trained model on the source domain and have to adapt it to the new incoming data without accessing the ground truth annotation. Herein, we propose a framework of complementary label learning with a consistent risk, which enables the model of its learning objective to agree with minimizing the real classification risk in the unseen dataset.

\paragraph{\bf Assigning the pseudo labels}
The most challenging problem of TTA is how to represent the labels of instances in the target domain. 
Obviously, the most direct way to produce pseudo labels is to assign the most probable class by reusing the knowledge $f^{\text{src}}(\cdot)$ learned from the source domain. 
The decision function $f^{\text{src}}(x_{i}): \mathcal{X} \rightarrow \mathbb{R}^{K}$ is parameterized by the output layer of a deep network where $K$ is the number of classes.
The decision function suggests a temporary appropriate data distribution $\Hat{p}^{*}(y|x=x_{i}) = f^{\text{src}}(x_{i})$. For the ordinary one-versus-zero label distribution, the general pseudo label of instance $x_{i}$ is 
\begin{equation}
    \left\{
    \begin{array}{lc}
         1 &  \text{if} \quad y=\mathop{\arg\max}\limits_{k}([f^{\text{src}}(x_{i})_{k}]), \\
         0 &  \text{else} ,
    \end{array}
    \right.
\end{equation}

\par 
where $[]_{k}$ denotes the $k$-th element. However, this implementation of pseudo labels requires enough confidence in decision function $f^{\text{src}}(\cdot)$. If fed with too much incorrect information, the model's performance may be severely damaged. The one-versus-zero label distribution seems to be greatly demanding, resulting in the model easily suffering extremely sensitive errors. Herein, we propose a complementary label density to formulate the soft complementary labels (CLs). Similarly, we reuse the decision classifier $f^{\text{src}}$ as a prior label distribution $\hat{p}^{*}$ and then filter it with a predefined threshold to obtain the distribution of complementary labels.
\begin{equation}\label{eq: CLs}
    \bar{p}^{*}(\bar{y}|x=x_{i})  =\dfrac{\left[\theta - f^{\text{src}}(x_{i})\right]^{T}_{+}}{\theta^{T}e_{\bar{y}}} \cdot e_{\bar{y}},
\end{equation}
where $[\cdot]_{+}$ represents the operator that remains the positive part else zero; $\theta$ denotes the maximum probability to filter the complementary labels and $e_{i}$ the $i$-th unit vector. Note that we have ignored the negative probability of complementary labels for avoiding polluting the model and scale it with its corresponding threshold of each class.
\begin{assumption}
\label{assumption}
We assume the conditional distribution $\hat{p}^{*}(y|x)$ is sufficiently close to the true distribution $p^{*}(y|x) $in their functional space. It equally says that, learning with the distribution $\hat{p}^{*}$ in the target domain makes the corresponding classifier $f^{\text{target}}$ return a similar performance on the data with the true distribution.
\end{assumption}
Assumption~\ref{assumption} simply offers us a view that starting with prior estimation $\hat{p}^{*}(y|x)=f^{\text{src}}(x)$ is feasible as Eq.~\ref{eq: CLs}. The conditional complementary label distribution is exactly reflecting the probability that this instance $x_{i}$ does not belong to the class $y_{i}$.

\paragraph{\bf Designing the complementary risk formula}
In the following, we will use the shorthand notation $\hat{p}^{*}$ and $\bar{p}^{*}$ to represent the data distribution $\hat{p}^{*}(y|x)$ and $\bar{p}^{*}(\bar{y}|x)$. In the target domain, the marginal distribution $p^{*}(x)$ is the same for the ordinary dataset and complementary dataset. So, we can rewrite Eq.~\ref{eq: CLs}
as:
\begin{equation}\label{conservation}
    \bar{p}^{*} = \dfrac{1}{\theta} \circ \left(\theta e^{T} - I \right) \hat{p}^{*},
\end{equation}
where $I$ is the unit matrix. Note that we have dropped the operator $[\cdot]_{+}$ in the derivation of Equ.~\ref{conservation} for a compact formula. 

\par
Consequently, 
\begin{equation}
\begin{aligned}
    \hat{p}^{*} &= -\left(I-\theta e^{T}\right)^{-1}\Theta\bar{p}^{*} \\
    &=-\left(I + \dfrac{1}{1 - e^{T}\theta}\theta e^{T}\right)\Theta \bar{p}^{*}, \\
\end{aligned}
\end{equation}
where $\Theta$ is diagonal matrix of vector $\theta$ and the second equality holds since the Sherman-Morrison formula. This conservation gives us an explicit relationship between estimated distribution $\hat{p}^{*}$ and our defined complementary distribution $\bar{p}^{*}$. 
\par 
Finally, we derive the risk function in complementary data distribution $\bar{p}^{*}$ to agree with the data defined in ordinary data distribution $\hat{p}^{*}$,
\begin{equation}
\begin{aligned}
     \hat{\mathcal{R}}(\theta)  &=\langle-log\, p_{\theta}(y|x)\rangle_{\hat{p}^{*}} \\
     &=e^{T} \left[-log\,p_{\theta}(y|x)\hat{p}^{*}(y|x)\right]e \\
     &=e^{T} \left[-log\,p_{\theta}(y|x)\left(-I - \frac{1}{1 - e^{T}\theta}\theta e^{T}\right)\Theta \bar{p}^{*} \right] e.
\end{aligned}
\end{equation}
where the conditional distribution $p_{\theta}(y|x)$ is the optimization objective and it can be simply represented as decision function $f^{\text{test}}$. It is easy to implement by incorporating the definition of the complementary label in Equ.~\ref{eq: CLs} as
\begin{equation}
    \label{eq: risk}
    \begin{aligned}
        &\bar{\mathcal{R}}(\theta)= \dfrac{1}{N}\sum_{x_{i}}\sum_{\bar{y}|\bar{p}^{*}>0}-\frac{[\theta_{\bar{y}^{*}}-f^{\text{src}}(x_i)_{\bar{y}}]_+}{\theta_{\bar{y}}}\\
    &\left[\theta_{\bar{y}}[logf^{\text{test}}(x_i)]_{\bar{y}} + \dfrac{1}{1-e^{T}\theta}\theta_{\bar{y}}\sum_{j}\theta_{i}[logf^{\text{test}}(x_i)]_{j}\right].
    \end{aligned}
\end{equation}
\par
We have recovered the original definition of Equ.~\ref{eq: CLs} in the second summation of Equ.~\ref{eq: risk}. This algorithm only computes the loss of valid negative categories, avoiding the influence of ambiguous and uncertain instances.

\begin{figure}
    \centering
    \includegraphics[width=1.\linewidth]{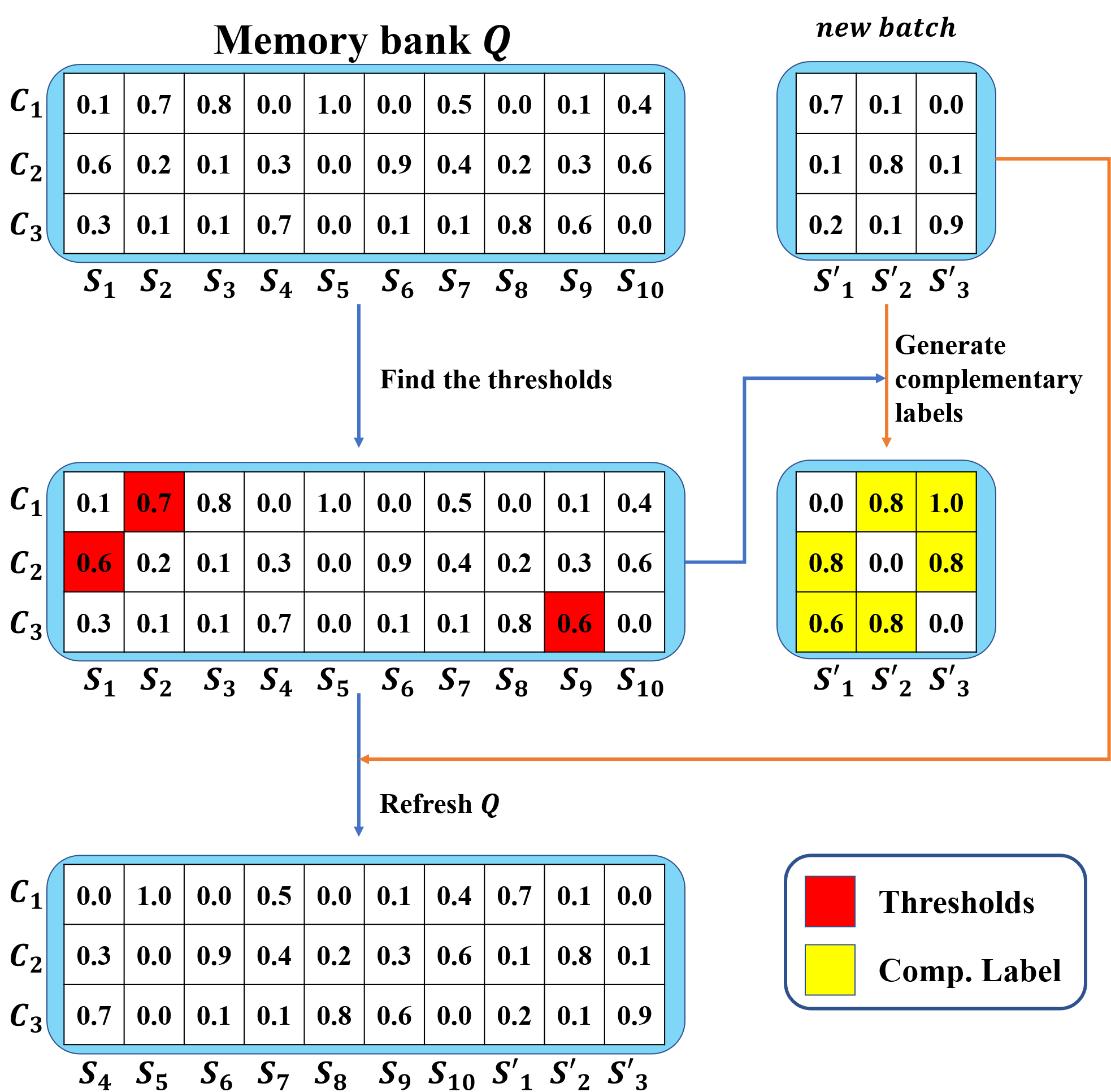}
    \caption{A brief illustration of the proposed dynamic thresholding strategy. For each category, we find its threshold according to $Q$. We then combine the thresholds and the prediction of each sample to generate their complementary labels to train the model. After that, predictions are utilized to refresh the memory bank.}
    \label{fig:queue}
\end{figure}

\paragraph{\bf Thresholding strategy.}
To find the proper threshold $\theta$ introduced in the past sections, we employ two thresholding techniques: a fixed thresholding strategy and a dynamic thresholding strategy. The fixed strategy involves treating the threshold as a hyperparameter and maintaining it constant across all categories during the TTA procedure. The dynamic strategy, on the other hand, adapts the threshold during the procedure in response to the probability distribution of past samples.
To adapt the threshold during testing, we propose to store the output distributions in a memory bank, which is denoted as $Q$. We utilize the probability distribution of $Q$ as the prior distribution of $p(y)$. The threshold of the $i_{th}$ category $\theta_i$ is calculated as follows:
\begin{equation}
    \begin{array}{l}
        \theta_i = \operatorname{Percentile}(Q_{i,:}, t),
    \end{array}
\end{equation}
in which $t$ is a hyperparameter. To a new batch of data, denote the output matrix as $D\in [0,1]^{N\times C}$ in which $N$ represents the number of samples, and $C$ represents the number of categories. We calculate the complementary label matrix $\Tilde{D}$ as in Equ.~\ref{equ:hcl}:
\begin{equation}
    \begin{array}{l}
        \Tilde{D}_{i,j} = \frac{[\operatorname{Percentile}(Q_{j,:}, t)-D_{i,j}]_+}{\operatorname{Percentile}(Q_{j,:}, t)}.
    \end{array}
    \label{equ:hcl}
\end{equation}
After each batch, we refresh $Q$ via the current batch of data. First, we merge $Q$ and $D$ together. If $\vert Q \vert > L $ in which $L$ is the largest number of samples to be stored in $Q$, we delete the first $\vert Q \vert - L$ columns to meet $\vert Q\vert \leq L$. $\vert Q \vert$ represents the number of samples stored in $Q$ currently.
A brief illustration of this procedure is shown in Fig.~\ref{fig:queue}. 
\section{Experiments}

\subsection{BCL with Known Complementary Label.}
We conduct an experiment to prove that the proposed complementary learning is capable of training a classifier. We train a toy model with 5 convolution layers and 2 MLPs to classify the CIFAR-10 dataset. The baseline is trained with standard ground truth labels. We then turn to BCL, and gradually decrease the number of given negative categories $N$ for each input sample. During the training process, the complementary labels are fixed. The result is shown in Tab.~\ref{tab:demo}. The performance of BCL is strongly related to $N$ because a small $N$ results in a large uncertainty of the ground truth positive label. However, the model trained with BCL is still capable of learning an effective classifier.

\begin{table}[htbp]
    \centering
    \begin{tabular}{c|c|c|c}
    \toprule
         N=4    & N = 6    & N = 8    & baseline  \\
    \midrule
         64.9\%        & 73.5\%     & 78.4\%     & 81.7\% \\
    \bottomrule
    \end{tabular}
    \caption{BCL on CIFAR-10 with known negative labels. $N$ represents how many negative labels of each sample are given.}
    \label{tab:demo}
\end{table}

\subsection{TTA Implementation Details}
We follow the corresponding pre-processing and utilize the pre-trained model as in website\footnote{https://github.com/huyvnphan/PyTorch\_CIFAR10} for CIFAR-10 and website\footnote{https://github.com/weiaicunzai/pytorch-cifar100} for CIFAR-100.
We utilize ResNet-18 for the CIFAR-10, and ResNet-50 for the CIFAR-100, respectively. In the following experiments, we follow the settings in \cite{wang2020tent} and \cite{wang2022continual}. We set the maximum length $L$ of $Q$ to $200$ and the percentile $p$ to $75\%$. We follow \cite{wang2020tent} to set the learning rate to 1e-3 for CIFAR-100-C. For CIFAR-10-C, the learning rate is set to 1e-4.

\subsection{Comparison with SOTA Approaches}
To evaluate the proposed complementary learning, we try two different experiment settings. For the ``one at a time (OAAT)'' setting, we assume that we only need to face one corruption type. Therefore, after finetuning the model on each corruption type with severity 5, we refresh the model with the initial pre-trained model on CIFAR-10 and CIFAR-100, respectively. We show the results in Tab.~\ref{tab:CIFAR10} and \ref{tab:CIFAR100}. For the ``Continual'' setting, we assume that the corruption type is changing with time. In this setting, we don't refresh the model. We show the results in Tab.~\ref{tab:CIFAR10con} and \ref{tab:CIFAR100con}. ``Mean'' represents the average performance under the involved corruption types. We bold the best performance under each type of corruption. It is worth noting that we choose the best thresholding strategy for BCL and ECL. A detailed comparison between the strategies could be found in the ablation study. The proposed approach achieves the best performance on both datasets and experiment settings, which demonstrates the effectiveness and robustness of the proposed complementary learning.

\begin{table*}[htbp]
    \centering
    \resizebox{\textwidth}{30mm}{
      \begin{tabular}{ccccccccccc}
      \hline
      Model & contrast & gau\_noise & impulse & brightness & saturate & glass & defocus & spatter & speckle & elastic \\
      \hline
      Source & 19.77  & 36.87  & 20.65  & 87.98  & 84.63  & 57.64  & 65.41  & 71.33  & 48.14  & 77.33  \\
      BN    & 81.43  & 71.03  & 53.07  & 89.08  & 89.01  & 64.39  & 85.26  & 75.24  & 68.82  & 77.74  \\
      Tent\cite{wang2020tent}  & 84.75  & 76.50  & 60.04  & 90.34  & 90.15  & 69.38  & {\bf 87.34}  & 79.52  & 77.17  & 80.30  \\
      CoTTA\cite{wang2022continual} & {\bf 86.28}  & 76.88  & {\bf 63.72}  & 90.10  & 89.87  & 67.85  & 87.13  & 79.58  & 78.15  & 81.07  \\
      {\bf BCL(ours)} & 85.96  & 77.86  & 62.55  & 90.31  & 90.14  & 71.74  & 87.22  & 80.58  & 78.77  & 81.04  \\
      {\bf ECL(ours)} & 85.68  & {\bf 79.19}  & 63.71  & {\bf 90.39}  & {\bf 90.38}  & {\bf 72.52}  & 87.24  & {\bf 80.98}  & {\bf 79.84}  & {\bf 81.22}  \\
      \midrule
      Model & pixelate & fog   & jpeg  & motion & snow  & frost & shot  & zoom  & gau\_blur & {\bf \underline{Mean}} \\
      \hline
      Source & 69.34  & 65.35  & 80.52  & 69.17  & 79.99  & 71.33  & 45.50  & 78.24  & 58.66  & 62.52  \\
      BN    & 81.08  & 81.48  & 79.05  & 84.22  & 82.20  & 79.61  & 72.39  & 85.26  & 84.94  & 78.17  \\
      Tent\cite{wang2020tent}  & 85.35  & 85.57  & 81.51  & 85.73  & 84.80  & 82.32  & 79.50  & {\bf 88.06}  & 86.80  & 81.85  \\
      CoTTA\cite{wang2022continual} & 84.93  & 86.15  & 82.25  & 86.10  & 85.28  & 83.45  & 79.97  & 87.64  & {\bf 87.19}  & 82.29  \\
      {\bf BCL(ours)} & 85.71  & 86.27  & 82.12  & {\bf 86.16}  & 85.23  & 83.65  & 80.78  & 87.83  & 86.88  & 82.67  \\
      {\bf ECL(ours)} & {\bf 86.04}  & {\bf 86.86}  & {\bf 82.30}  & 86.06  & {\bf 85.51}  & {\bf 83.93}  & {\bf 81.58}  & 88.05  & 87.13  & {\bf \underline{83.12}}  \\
      \bottomrule
      \end{tabular}
      }%
      
      \caption{Comparison with SOTA approaches of OAAT setting on CIFAR-10-C.}
    \label{tab:CIFAR10}%
  \end{table*}%

  \begin{table*}[htbp]
    \centering
    \resizebox{\textwidth}{30mm}{
      \begin{tabular}{ccccccccccc}
      \toprule
      Model & contrast & gau\_noise & impulse & brightness & saturate & glass & defocus & spatter & speckle & elastic \\
      \hline
      Source & 16.28 & 12.72 & 7.15  & 65.23 & 56.37 & 46.90  & 33.81 & 60.85 & 15.66 & 57.91 \\
      BN    & 63.18 & 41.05 & 39.63 & 66.94 & 66.05 & 52.97 & 66.01 & 64.36 & 40.90  & 60.04 \\
      Tent\cite{wang2020tent}  & 66.07 & 53.26 & 51.32 & 71.92 & 69.87 & 56.82 & 69.03 & 72.16 & 53.43 & 63.77 \\
      CoTTA\cite{wang2022continual} & 63.72 & 41.31 & 40.13 & 67.44 & 66.09 & 53.51 & 66.62 & 65.02 & 40.96 & 60.84 \\
      {\bf BCL(ours)} & {\bf 69.76} & 55.33 & 54.19 & 73.08 & {\bf 72.26} & 61.02 & {\bf 71.87} & 73.35 & 57.32 & 65.49 \\
      {\bf ECL(ours)} & 69.67 & {\bf 55.58} & {\bf 54.32} & {\bf 73.79} & 72.17 & {\bf 61.27} & 71.81 & {\bf 73.54} & {\bf 57.85} & {\bf 65.68} \\
      \midrule
      Model & pixelate & fog   & jpeg  & motion & snow  & frost & shot  & zoom  & gau\_blur & {\bf \underline{Mean}}\\
      \hline
      Source & 33.42 & 38.28 & 48.20  & 45.08 & 50.70  & 43.11 & 14.57 & 45.17 & 26.18 &  37.77\\
      BN    & 61.69 & 56.76 & 49.23 & 64.24 & 55.17 & 55.77 & 41.96 & 66.47 & 64.52 &  56.68\\
      Tent\cite{wang2020tent}  & 67.28 & 64.69 & 56.30  & 67.60  & 63.12 & 62.26 & 55.56 & 69.75 & 69.17 & 63.34 \\
      CoTTA\cite{wang2022continual} & 62.04 & 56.87 & 49.22 & 64.39 & 55.36 & 56.23 & 42.35 & 67.31 & 65.19 &  57.08\\
      {\bf BCL(ours)} & {\bf 69.56} & {\bf 66.66} & {\bf 58.69} & 69.73 & 65.03 & 64.47 & 57.59 & 71.46 & {\bf 71.95} & 65.73 \\
      {\bf ECL(ours)} & 68.91 & 66.61 & 58.22 & {\bf 69.76} & {\bf 65.60} & {\bf 64.80} & {\bf 58.06} & {\bf 71.61} & 71.36 & {\bf \underline{65.82}} \\
      \bottomrule
      \end{tabular}
      }%
      \caption{Comparison with SOTA approaches of OAAT setting on CIFAR-100-C.}
    \label{tab:CIFAR100}%
  \end{table*}%

  \begin{table*}[htbp]
    \centering
    \resizebox{\textwidth}{30mm}{
      \begin{tabular}{ccccccccccc}
      \toprule
      Model & saturate & gau\_blur & glass & defocus & spatter & speckle & elastic & pixelate & contrast & gau\_noise \\
      \hline
      Order  & \multicolumn{9}{c}{$\xrightarrow{\quad\quad\quad\quad\quad\quad\quad\quad\quad\quad\quad\quad\quad\quad\quad\quad\quad\quad\quad\quad}$} &  \\
      \hline
      Source & 84.63  & 58.66  & 57.64  & 65.41  & 71.33  & 48.14  & 77.33  & 69.34  & 19.77  & 36.87  \\
      BN    & 89.01  & 84.94  & 64.39  & 85.26  & 75.24  & 68.82  & 77.74  & 81.08  & 81.43  & 71.03  \\
      Tent\cite{wang2020tent}  & 89.95  & 86.34  & 68.66  & {\bf 86.44}  & 77.04  & 72.66  & 77.80  & 82.41  & 77.74  & 71.28  \\
      CoTTA\cite{wang2022continual} & 88.08  & 85.63  & {\bf 71.40}  & 82.83  & 76.93  & 74.53  & 75.79  & 79.94  & {\bf 82.50}  & 74.52  \\
      {\bf BCL(ours)} & 90.04  & 86.94  & 70.65  & 85.82  & {\bf 78.81}  & 75.68  & 79.92  & 83.89  & 81.61  & 76.00  \\
      {\bf ECL(ours)} & {\bf 90.05}  & {\bf 86.97}  & 70.57  & 86.42  & 78.65  & {\bf 76.01}  & {\bf 80.22}  & {\bf 84.25}  & 81.52  & {\bf 76.34}  \\
      \midrule
      Model & zoom  & shot  & impulse & fog   & frost & snow  & motion & jpeg  & brightness & {\bf \underline{Mean}} \\
      \hline
      Order  & \multicolumn{9}{c}{$\xrightarrow{\quad\quad\quad\quad\quad\quad\quad\quad\quad\quad\quad\quad\quad\quad\quad\quad\quad\quad\quad\quad}$} &  \\
      \hline
      Source & 78.24  & 45.50  & 20.65  & 65.35  & 71.33  & 79.99  & 69.17  & 80.52  & 87.98  & 62.51  \\
      BN    & 85.26  & 72.39  & 53.07  & {\bf 81.48}  & 79.61  & {\bf 82.20}  & {\bf 84.22}  & 79.05  & {\bf 89.08}  & 78.17  \\
      Tent\cite{wang2020tent}  & 82.22  & 71.29  & 54.19  & 75.65  & 74.43  & 75.06  & 75.71  & 73.80  & 81.47  & 76.53  \\
      CoTTA\cite{wang2022continual} & 81.73  & 74.72  & {\bf 62.96}  & 69.74  & 69.95  & 70.89  & 71.93  & 66.15  & 69.11  & 75.23  \\
      {\bf BCL(ours)} & 86.10  & 78.90  & 62.41  & 80.13  & {\bf 80.42}  & 81.69  & 82.53  & 80.43  & 87.62  & 80.50  \\
      {\bf ECL(ours)} & {\bf 86.21}  & {\bf 79.03}  & 62.50  & 79.85  & 80.30  & 82.06  & 82.58  & {\bf 80.66}  & 87.88  & {\bf \underline{80.63}}  \\
      \bottomrule
      \end{tabular}
      }%
      \caption{Continual test-time adaptation on CIFAR-10-C.}
    \label{tab:CIFAR10con}%
  \end{table*}%

\begin{table*}[htbp]
    \centering
    \resizebox{\textwidth}{30mm}{
      \begin{tabular}{ccccccccccr}
      \toprule
      Model & saturate & gau\_blur & glass & defocus & spatter & speckle & elastic & pixelate & contrast & \multicolumn{1}{c}{gau\_noise} \\
      \hline
      Order  & \multicolumn{9}{c}{$\xrightarrow{\quad\quad\quad\quad\quad\quad\quad\quad\quad\quad\quad\quad\quad\quad\quad\quad\quad\quad\quad\quad}$} &  \\
      \hline
      Source & 56.37 & 26.18 & 46.90  & 33.81 & 60.85 & 15.66 & 57.91 & 33.42 & 16.28 & \multicolumn{1}{c}{12.72} \\
      BN    & 66.05 & 64.52 & 52.97 & 66.01 & 64.36 & 40.90  & 60.04 & 61.69 & 63.18 & \multicolumn{1}{c}{41.05} \\
      Tent\cite{wang2020tent}  & {\bf 71.98}  & 70.99  & 59.90  & 71.33  & 70.80  & 53.32  & 62.68  & 65.93  & 63.36  & \multicolumn{1}{c}{51.27} \\
      CoTTA\cite{wang2022continual} & 66.19 & 65.20  & 53.14 & 66.48 & 64.95 & 41.25 & 60.49 & 61.98 & 63.51 & \multicolumn{1}{c}{41.63} \\
      {\bf BCL(ours)} & 71.09  & {\bf 71.04}  & 60.25  & 70.20  & 69.23  & 56.30  & 63.24  & 66.09  & 61.30   & \multicolumn{1}{c}{54.21}\\
      {\bf ECL(ours)} & 70.26  & 69.93  & {\bf 61.35}  & {\bf 72.17}  & {\bf 71.59}  & {\bf 57.85}  & {\bf 66.45}  & {\bf 69.09}  & {\bf 66.07}   & \multicolumn{1}{c}{{\bf 56.72}}\\
      \midrule
      Model & zoom  & shot  & impulse & fog   & frost & snow  & motion & jpeg  & brightness &  \multicolumn{1}{c}{{\bf \underline{Mean}}}\\
      \hline
      Order  & \multicolumn{9}{c}{$\xrightarrow{\quad\quad\quad\quad\quad\quad\quad\quad\quad\quad\quad\quad\quad\quad\quad\quad\quad\quad\quad\quad}$} &  \\
      \hline
      Source & 45.17 & 14.57 & 7.15  & 38.28 & 43.11 & 50.70  & 45.08 & 48.20  & 65.23 & \multicolumn{1}{c}{37.76} \\
      BN    & 66.47 & 41.96 & 39.63 & 56.76 & 55.77 & 55.17 & 64.24 & 49.23 & 66.94 & \multicolumn{1}{c}{56.68} \\
      Tent\cite{wang2020tent}  & 65.50  & 51.88  & 44.60  & 51.20  & 52.73  & 50.22  & 54.30  & 45.72  & 55.52  & \multicolumn{1}{c}{58.59}  \\
      CoTTA\cite{wang2022continual} & 67.72 & 42.47 & 40.37 & 57.18 & 56.68 & 56.04 & 64.67 & 50.14 & 68.03 & \multicolumn{1}{c}{57.26} \\
      {\bf BCL(ours)} & 66.63  & 56.57  & 50.05  & 55.05  & 58.94  & 57.06  & 61.36  & 54.55  & 64.96  & \multicolumn{1}{c}{61.48}  \\
      {\bf ECL(ours)} & {\bf 70.90}  & {\bf 59.91}  & {\bf 54.17}  & {\bf 60.30}  & {\bf 63.67}  & {\bf 61.25}  & {\bf 66.12}  & {\bf 58.48}  & {\bf 69.68}  & \multicolumn{1}{c}{{\bf \underline{64.52}}}  \\
      \bottomrule
      \end{tabular}
      }%
      \caption{Continual test-time adaptation on CIFAR-100-C.}
    \label{tab:CIFAR100con}%
  \end{table*}%

\subsection{Ablation Study}
\paragraph{The order of corruption types.} To evaluate the influence of corruption types, we randomly shuffle the corruption types 5 times and calculate the average performance on CIFAR-10-C. Note that we do not include the result in Tab.~\ref{tab:CIFAR10con}. The result is shown in Tab.~\ref{tab:shuffle}. The proposed complementary learning shows the consistency of different corruption orders. 

\begin{table}[H]
  \centering
  
    \begin{tabular}{c|c|c|c}
    \toprule
    Shuffle Idx & 1     & 2     & 3      \\
    Perf. (\%)  & 80.87 & 80.92 & 81.38  \\
    \midrule
    Shuffle Idx & 4     & 5     & Mean \\
    Perf. (\%)  & 80.87 & 81.15 & 81.04±0.20 \\
    \bottomrule
    \end{tabular}%
    \caption{Performance of CL on shuffled CIFAR-10-C under ``Continual'' setting.}
  \label{tab:shuffle}%
\end{table}%

\paragraph{Trained parameters.} In all the above experiments, we follow Tent \cite{wang2020tent} to only finetune the parameters of batch normalization layers. Here we conduct an experiment on the trained parameters in TTA with BCL. We show the result in Tab.~\ref{tab:para}. "BN" finetunes batch normalization parameters only, "Feature" finetunes all model parameters except final FC layers, "Classifier" finetunes final FC layers, and "All" finetunes the entire model. Finetune the overall model has similar performance compared with only finetuning the BN layers. However, only finetuning the feature extractor or FC layers achieves much worse performance.

\begin{table}[H]
    \centering
    \begin{tabular}{c|c|c|c|c}
    \toprule
         Param     & BN    & Feature   & Classifier   & All    \\
         \midrule
         Perf.(\%) & 82.67 & 78.29     & 78.46        & 82.68  \\
    \bottomrule
    \end{tabular}
    \caption{Performance of BCL with different finetuned parameter groups on CIFAR-10-C under OAAT setting: "BN" finetunes batch normalization parameters only, "Feature" finetunes all model parameters except final FC layers, "Classifier" finetunes final FC layers, and "All" finetunes the entire model.}
    \label{tab:para}
\end{table}

\paragraph{\bf The maximum length of the memory bank $Q$.} We conduct an experiment to search for the best length $L$ on all datasets and experiment settings. The result is shown in Tab.~\ref{tab:L}. Increasing $L$ can slightly improve the performance of BCL generally since it provides a better distribution evaluation of each category. Detailed results for ECL are available in the supplementary material. However, since larger $L$ takes longer time during training, we empirically set it to 200.

\begin{table}[H]
    \centering
    \begin{tabular}{c|c|c|c|c}
    \toprule
         Dataset    & Setting     & $L$=50        & $L$=200    &$L$=500    \\
         \midrule
         CIFAR-10-C    & OAAT        & 82.63           &  82.67     & 82.70          \\
         CIFAR-10-C    & Continual   & 79.95           &  80.50     & 80.57          \\
         \midrule
         CIFAR-100-C   & OAAT        & 65.55           &  65.73     & 65.76     \\
         CIFAR-100-C   & Continual   & 61.39           &  61.48     & 61.56     \\
    \bottomrule
    \end{tabular}
    \caption{Search the maximum length for $L$ on different datasets and experiment settings.}
    \label{tab:L}
\end{table}

\paragraph{\bf Comparison with naive pseudo-labeling.} We evaluate both ECL and naive pseudo-labeling (NPL) on CIFAR-10-C and CIFAR-100-C under both experiment settings. The result is shown in Tab.~\ref{tab:NPL}. The proposed ECL outperforms NPL in both settings, which suggests that ECL provides a more robust pseudo-supervision for the model to learn from.

\begin{table}[H]
    \centering
    \begin{tabular}{c|c|c}
    \toprule
         Model    & Setting     & Perf (\%)  \\
         \midrule
         NPL      & OAAT        & 70.81          \\
         ECL     & OAAT        & 74.20           \\
         \midrule
         NPL      & Continual   & 69.38           \\
         ECL     & Continual   & 72.56           \\
    \bottomrule
    \end{tabular}
    \caption{Comparison between ECL and NPL on both datasets and settings. Performance is the average of CIFAR-10-C and CIFAR-100-C.}
    \label{tab:NPL}
\end{table}

\paragraph{\bf Comparison of thresholding approaches.}
To validate the proposed thresholding approach, we compare it with fixed thresholds. For CIFAR-10 and CIFAR-100, we set the fixed threshold as $5e-2$ and $5e-3$, respectively. The result is shown in Tab.~\ref{tab:thres}. The result shows that the proposed thresholding works well both under the OAAT setting and the Continual setting, while the fixed thresholding failed for continual settings. This result shows that the dynamic threshold helps the model to optimize smoothly. ``Failed'' represents under the ``Continual'' setting, the model is unable to correctly predict the categories of the samples in later types of corruptions.

\begin{table}[H]
  \centering
  
    \begin{tabular}{c|c|c|c|c}
    \toprule
    Dataset      & \multicolumn{2}{c|}{CIFAR-10-C} & \multicolumn{2}{c}{CIFAR-100-C} \\
    \midrule
    Setting      & Continual & OAAT  & Continual & OAAT \\
    \midrule
    BCL-fixed & 45.66  & 82.96  & failed   & 63.69  \\
    BCL-dynamic & 80.50  & 82.67  & 61.48  & 65.73  \\
    ECL-fixed & 79.76 & 83.12  & failed & 65.82  \\
    ECL-dynamic & 80.63  & 82.08  & 64.52  & 65.42  \\
    \bottomrule
    \end{tabular}%
    \caption{Comparison between the proposed thresholding (dynamic) and fixed thresholding.}
  \label{tab:thres}%
\end{table}%

\begin{figure}[htbp]
    \centering
    \includegraphics[width=.95\linewidth]{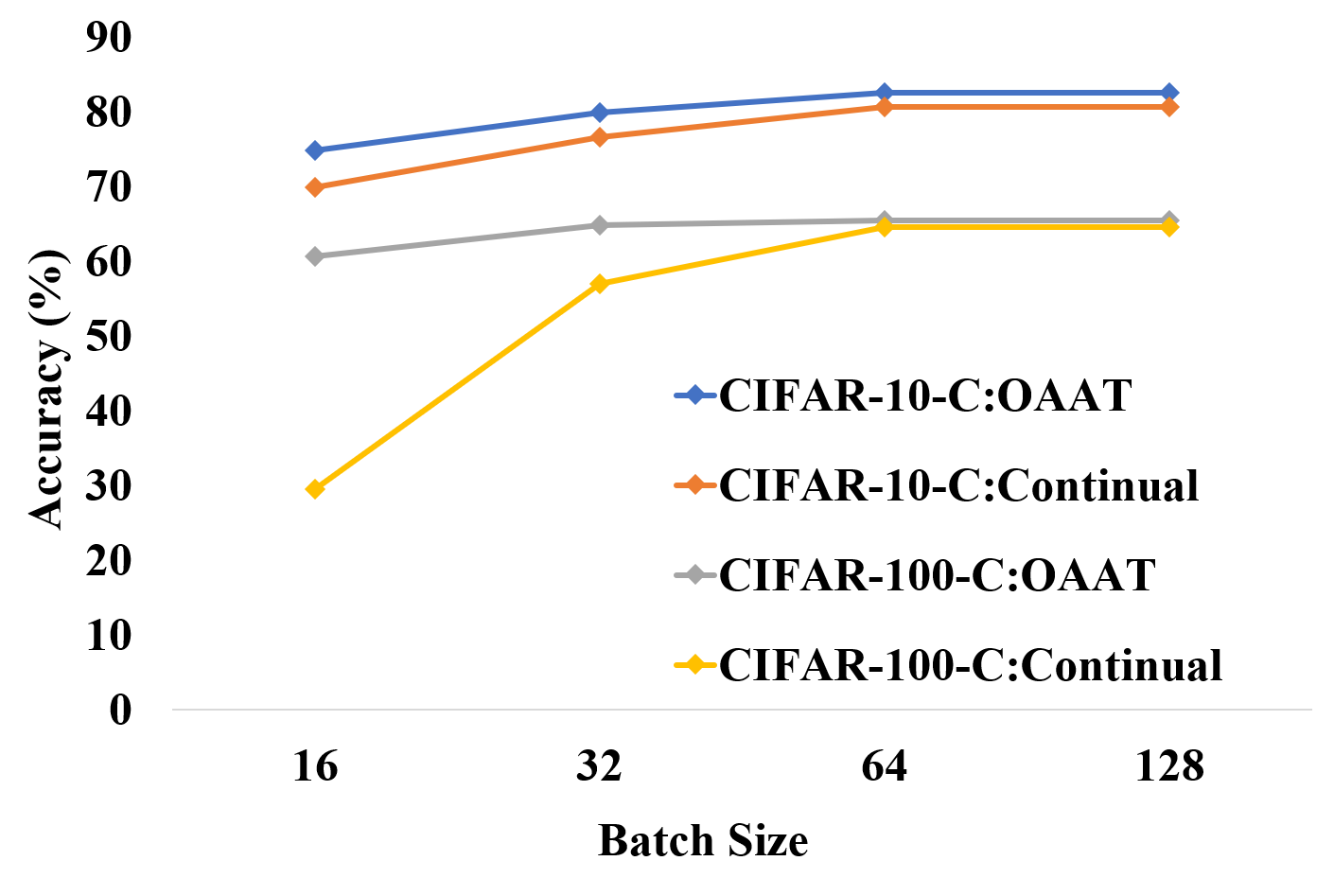}
    \caption{The effect of different batch sizes on ECL with dynamic thresholding. }
    \label{fig:bs}
\end{figure}

\paragraph{\bf Validation of batch size.}
We further validate the effect of different batch sizes conditioned on ECL with dynamic thresholding, and the result is shown in Fig.~\ref{fig:bs}. When the batch size is small, the TTA performance could be significantly affected by it. When the batch size is larger than $64$, it becomes less important to the final performance.
\section{Conclusion}
In this work, we propose a TTA framework, named complementary learning, that explicitly takes the advantage of the non-maximal categories of the input samples. We start with descriptions of complementary labels and why are complementary labels more accurate than pseudo labels. Then we give a basic but effective approach to utilizing the complementary labels. We further introduce confidence to the basic formulation to give a more effective one. To dynamically adapt the threshold, we propose a simple queue-based approach. 
Experiments on different experiment settings (continual and OAAT) and datasets (CIFAR-10 and CIFAR-100) show that the proposed approach achieves state-of-the-art performance on test-time adaptation tasks.

\bibliographystyle{named}
\bibliography{ref}

\begin{thebibliography}{}

\bibitem[\protect\citeauthoryear{Chen \bgroup \em et al.\egroup
  }{2019}]{chen2019joint}
Chao Chen, Zhihong Chen, Boyuan Jiang, and Xinyu Jin.
\newblock Joint domain alignment and discriminative feature learning for
  unsupervised deep domain adaptation.
\newblock In {\em Proceedings of the AAAI conference on artificial
  intelligence}, volume~33, pages 3296--3303, 2019.

\bibitem[\protect\citeauthoryear{Chen \bgroup \em et al.\egroup
  }{2021}]{chen2021beyond}
Pengfei Chen, Junjie Ye, Guangyong Chen, Jingwei Zhao, and Pheng-Ann Heng.
\newblock Beyond class-conditional assumption: A primary attempt to combat
  instance-dependent label noise.
\newblock In {\em Proceedings of the AAAI Conference on Artificial
  Intelligence}, volume~35, pages 11442--11450, 2021.

\bibitem[\protect\citeauthoryear{Kim \bgroup \em et al.\egroup
  }{2021}]{kim2021fine}
Taehyeon Kim, Jongwoo Ko, JinHwan Choi, Se-Young Yun, et~al.
\newblock Fine samples for learning with noisy labels.
\newblock {\em Advances in Neural Information Processing Systems},
  34:24137--24149, 2021.

\bibitem[\protect\citeauthoryear{Liang \bgroup \em et al.\egroup
  }{2020}]{liang2020we}
Jian Liang, Dapeng Hu, and Jiashi Feng.
\newblock Do we really need to access the source data? source hypothesis
  transfer for unsupervised domain adaptation.
\newblock In {\em International Conference on Machine Learning}, pages
  6028--6039. PMLR, 2020.

\bibitem[\protect\citeauthoryear{Liu \bgroup \em et al.\egroup
  }{2018}]{liu2018detach}
Yen-Cheng Liu, Yu-Ying Yeh, Tzu-Chien Fu, Sheng-De Wang, Wei-Chen Chiu, and
  Yu-Chiang~Frank Wang.
\newblock Detach and adapt: Learning cross-domain disentangled deep
  representation.
\newblock In {\em Proceedings of the IEEE Conference on Computer Vision and
  Pattern Recognition}, pages 8867--8876, 2018.

\bibitem[\protect\citeauthoryear{Long \bgroup \em et al.\egroup
  }{2013}]{Long_2013_ICCV}
Mingsheng Long, Jianmin Wang, Guiguang Ding, Jiaguang Sun, and Philip~S. Yu.
\newblock Transfer feature learning with joint distribution adaptation.
\newblock In {\em Proceedings of the IEEE International Conference on Computer
  Vision (ICCV)}, December 2013.

\bibitem[\protect\citeauthoryear{Mintun \bgroup \em et al.\egroup
  }{2021}]{mintun2021interaction}
Eric Mintun, Alexander Kirillov, and Saining Xie.
\newblock On interaction between augmentations and corruptions in natural
  corruption robustness.
\newblock {\em Advances in Neural Information Processing Systems},
  34:3571--3583, 2021.

\bibitem[\protect\citeauthoryear{Mummadi \bgroup \em et al.\egroup
  }{2021}]{mummadi2021test}
Chaithanya~Kumar Mummadi, Robin Hutmacher, Kilian Rambach, Evgeny Levinkov,
  Thomas Brox, and Jan~Hendrik Metzen.
\newblock Test-time adaptation to distribution shift by confidence maximization
  and input transformation.
\newblock {\em arXiv preprint arXiv:2106.14999}, 2021.

\bibitem[\protect\citeauthoryear{Wang \bgroup \em et al.\egroup
  }{2020}]{wang2020suppressing}
Kai Wang, Xiaojiang Peng, Jianfei Yang, Shijian Lu, and Yu~Qiao.
\newblock Suppressing uncertainties for large-scale facial expression
  recognition.
\newblock In {\em Proceedings of the IEEE/CVF Conference on Computer Vision and
  Pattern Recognition}, pages 6897--6906, 2020.

\bibitem[\protect\citeauthoryear{Wang \bgroup \em et al.\egroup
  }{2021}]{wang2020tent}
Dequan Wang, Evan Shelhamer, Shaoteng Liu, Bruno Olshausen, and Trevor Darrell.
\newblock Tent: Fully test-time adaptation by entropy minimization.
\newblock In {\em International Conference on Learning Representations (ICLR)},
  2021.

\bibitem[\protect\citeauthoryear{Wang \bgroup \em et al.\egroup
  }{2022}]{wang2022continual}
Qin Wang, Olga Fink, Luc Van~Gool, and Dengxin Dai.
\newblock Continual test-time domain adaptation.
\newblock In {\em Proceedings of the IEEE/CVF Conference on Computer Vision and
  Pattern Recognition}, pages 7201--7211, 2022.

\bibitem[\protect\citeauthoryear{Wu \bgroup \em et al.\egroup
  }{2020}]{wu2020topological}
Pengxiang Wu, Songzhu Zheng, Mayank Goswami, Dimitris Metaxas, and Chao Chen.
\newblock A topological filter for learning with label noise.
\newblock {\em Advances in neural information processing systems},
  33:21382--21393, 2020.

\bibitem[\protect\citeauthoryear{Xia \bgroup \em et al.\egroup
  }{2021}]{xia2021sample}
Xiaobo Xia, Tongliang Liu, Bo~Han, Mingming Gong, Jun Yu, Gang Niu, and Masashi
  Sugiyama.
\newblock Sample selection with uncertainty of losses for learning with noisy
  labels.
\newblock {\em arXiv preprint arXiv:2106.00445}, 2021.

\bibitem[\protect\citeauthoryear{Yu \bgroup \em et al.\egroup
  }{2018}]{yu2018learning}
Xiyu Yu, Tongliang Liu, Mingming Gong, and Dacheng Tao.
\newblock Learning with biased complementary labels.
\newblock In {\em Proceedings of the European conference on computer vision
  (ECCV)}, pages 68--83, 2018.

\bibitem[\protect\citeauthoryear{Zhang \bgroup \em et al.\egroup
  }{2021}]{zhang2021understanding}
Chiyuan Zhang, Samy Bengio, Moritz Hardt, Benjamin Recht, and Oriol Vinyals.
\newblock Understanding deep learning (still) requires rethinking
  generalization.
\newblock {\em Communications of the ACM}, 64(3):107--115, 2021.

\bibitem[\protect\citeauthoryear{Zheng \bgroup \em et al.\egroup
  }{2020}]{zheng2020error}
Songzhu Zheng, Pengxiang Wu, Aman Goswami, Mayank Goswami, Dimitris Metaxas,
  and Chao Chen.
\newblock Error-bounded correction of noisy labels.
\newblock In {\em International Conference on Machine Learning}, pages
  11447--11457. PMLR, 2020.

\end{thebibliography}

\end{document}